\documentclass[10pt,twocolumn,letterpaper]{article}
\usepackage[accsupp]{axessibility}  
\usepackage[pagenumbers]{cvpr}      

\usepackage{listings}
\usepackage{xcolor}
\usepackage{setspace}

\definecolor{codegreen}{rgb}{0,0.6,0}
\definecolor{codegray}{rgb}{0.5,0.5,0.5}
\definecolor{codeblue}{rgb}{0.0,0.0,0.8}
\definecolor{backcolour}{rgb}{0.98,0.98,0.98}

\lstdefinestyle{pytorchstyle_compact}{
    language=Python,
    backgroundcolor=\color{backcolour},   
    commentstyle=\color{codegreen},
    keywordstyle=\color{codeblue},
    stringstyle=\color{purple},
    basicstyle=\ttfamily\footnotesize\setstretch{1.1}, 
    breakatwhitespace=false,         
    breaklines=true,                 
    captionpos=b,                    
    keepspaces=true,                 
    numbers=none,                    
    numbersep=5pt,                  
    showspaces=false,                
    showstringspaces=false,
    showtabs=false,                  
    tabsize=2,
    frame=single, 
    rulecolor=\color{black},
    framesep=2mm,
    morekeywords={*,self,torch,nn,super,True,False,None,return,with,no_grad,F},
    escapechar=`
}

\makeatletter
\renewcommand\paragraph{%
  \@startsection{paragraph}{4}{\z@}%
    {0.4ex \@plus 0.5ex \@minus .2ex}
    {-0.5em}
    {\normalfont\normalsize\bfseries}%
}
\makeatother

\newcommand{\ours}{SL-HOI\xspace}

\newcommand{\dinonospace}{DINOv3}
\newcommand{\dino}{DINOv3\xspace}
\newcommand{\dinotxt}{\texttt{dino.txt}\xspace}
\newcommand{\cls}{\texttt{[CLS]}\xspace}

\newcommand{\hoi}{HOI\xspace}
\newcommand{\ov}{open-vocabulary\xspace}
\newcommand{\Ov}{Open-vocabulary\xspace}
\newcommand{\OV}{Open-Vocabulary\xspace}

\newcommand{\swig}{SWiG-HOI\xspace}
\newcommand{\hicodet}{HICO-DET\xspace}
\newcommand{\coco}{COCO\xspace}

\definecolor{cvprblue}{rgb}{0.21,0.49,0.74}
\usepackage[pagebackref,breaklinks,colorlinks,allcolors=cvprblue]{hyperref}

\def\paperID{}
\def\confName{CVPR}
\def\confYear{2026}

\title{Streamlined Open-Vocabulary Human-Object Interaction Detection}

\author{
Chang Sun \qquad Dongliang Liao \qquad Changxing Ding\thanks{Corresponding author} \\
South China University of Technology \\
{\tt\small eesunchang2024@mail.scut.edu.cn, \{liaodl, chxding\}@scut.edu.cn}
}

\begin{document}
\maketitle
\begin{abstract}
\quad \Ov human-object interaction (\hoi) detection aims to localize and recognize all human-object interactions in an image, including those unseen during training.
Existing approaches usually rely on the collaboration between a conventional \hoi detector and a Vision-Language Model (VLM) to recognize unseen \hoi categories.
However, feature fusion in this paradigm is challenging due to significant gaps in cross-model representations.
To address this issue, we introduce \textbf{\ours}, a \textbf{S}tream\textbf{L}ined open-vocabulary \textbf{\hoi} detection framework based solely on the powerful \dino model.
Our design leverages the complementary strengths of \dinonospace's components: its backbone for fine-grained localization and its text-aligned vision head for open-vocabulary interaction classification.
Moreover, to facilitate smooth cross-attention between the interaction queries and the vision head's output, we propose first feeding both the interaction queries and the backbone image tokens into the vision head, effectively bridging their representation gaps. All \dino parameters in our approach are frozen, with only a small number of learnable parameters added, allowing a fast adaptation to the \hoi detection task.
Extensive experiments show that \ours achieves state-of-the-art performance on both the \swig and \hicodet benchmarks, demonstrating the effectiveness of our streamlined model architecture.
Code is available at \href{https://github.com/MPI-Lab/SL-HOI}{https://github.com/MPI-Lab/SL-HOI}.
\end{abstract}
    
\section{Introduction}
\label{sec:intro}
Human-Object Interaction (\hoi) detection~\cite{VCOCO} is a fundamental vision task that involves not only localizing humans and objects in an image but also recognizing the interactions between each human-object pair.
It is critical for applications such as video analysis~\cite{VIDEO_ANALYSIS}, scene understanding~\cite{PGSG}, and robotics~\cite{ROBOTICS}.
Compared with object detection, \hoi detection is more dependent on the image context to infer the interaction categories.
Moreover, in the \ov setting, HOI detectors face the additional challenge of category generalization, requiring classification of long-tailed or even unseen HOI categories during training.

\begin{figure*}[ht]
    \centering
    \begin{subfigure}[b]{.31\linewidth}
        \includegraphics[width=\textwidth]{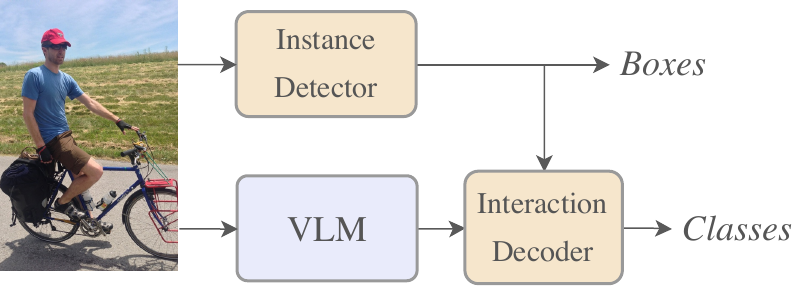}
        \caption{VLM-collaborated method.}
        \label{fig:first_method}
    \end{subfigure}
    \hfill
    \begin{subfigure}[b]{.24\linewidth}
        \includegraphics[width=\textwidth]{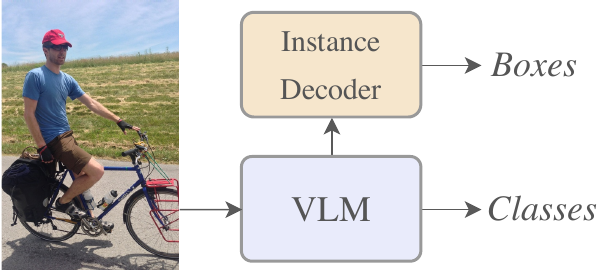}
        \caption{VLM-only method.}
        \label{fig:second_method}
    \end{subfigure}
    \hfill
    \begin{subfigure}[b]{.36\linewidth}
        \includegraphics[width=\textwidth]{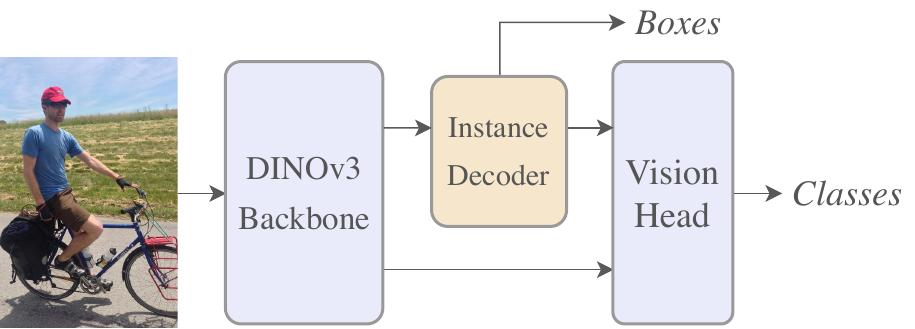}
        \caption{Our SL-HOI.}
        \label{fig:our_streamlined_method}
    \end{subfigure}
    \caption{An illustration of the dominant architectural paradigms for \ov \hoi detection. (a) VLM-collaborated methods that adopt both a VLM and a conventional \hoi detector. (b) VLM-only methods that employ a single VLM for \ov \hoi detection. (c) Our \ours leverages the complementary strengths of \dino's backbone and vision head.}
    \label{fig:methodology_comparisions}
\end{figure*}

Existing works rely on large-scale pre-trained Vision-Language Models (VLMs) to achieve \ov \hoi detection.
They can be categorized into two groups, as shown in Fig.~\ref{fig:first_method} and Fig.~\ref{fig:second_method}.
The first group of methods~\cite{GEN-VLKT,HOICLIP,UniHOI,BC-HOI} is based on collaboration between a VLM and a conventional \hoi detector, primarily by extracting generalizable interaction representations from the VLM for the \hoi detector.
The second group of approaches~\cite{THID,CMD-SE,SGC-Net,INP-CC} directly transforms a VLM into an \hoi detector for both interactive human-object detection and interaction classification. 

Unfortunately, both categories of methods have limitations.
Since the methods in the first group require two separately trained models, they tend to be complex in structure.
Moreover, the fusion of features between the \hoi detector and the VLM is challenging due to significant gaps in the cross-model representations.
The methods in the second category are generally based on the CLIP model~\cite{CLIP}. However, the CLIP model falls short in extracting fine-grained visual representations, since its training objective is to align holistic features between an image and its caption. The above analysis motivates us to develop the next-generation VLM-based \hoi detection model that is both simple in model structure and superior in \ov \hoi detection performance.

Accordingly, we propose a \textbf{S}tream\textbf{L}ined \ov \textbf{\hoi} detector, namely \textbf{\ours}, that streamlines interactive human-object detection and interaction classification.
Specifically, we adopt the \dinotxt variant~\cite{dinotxt} of the \dino model~\cite{DINOv3} as the VLM. This variant consists of a \dino backbone and a text-aligned vision head (hereafter “backbone” and “vision head”, respectively).
The backbone is pre-trained using large-scale self-supervised learning. 
It captures fine-grained visual features suitable for dense prediction, which we use for interactive human-object detection. 
To achieve this goal, we add a small detection decoder that uses the backbone's output patch tokens as the key and value. The vision head aligns visual features with \ov captions, which is ideal for generalizable interaction classification. 
Similar to popular one-stage detectors \hoi~\cite{GEN-VLKT}, the output embeddings of the detection decoder serve as interaction queries in this step. 

However, directly performing cross-attention between the interaction queries and the vision head's output still suffers from a representation gap. 
To address this problem, we propose to force the interaction queries and the vision head’s output tokens to share a common representation space. 
We achieve this by feeding both the interaction queries and the backbone's output image tokens into the vision head, rather than just the latter. 
Another advantage of this strategy is that it yields semantically enriched interaction queries. 
Then, we perform cross-attention between the refined interaction queries and the vision head’s output tokens, and the output is used for open-vocabulary interaction classification. 

In our approach, \dino serves as the sole backbone for \hoi detection, with all its parameters frozen. 
This streamlined design, as illustrated in Fig.~\ref{fig:our_streamlined_method}, contains only a small number of trainable parameters for an end-to-end \hoi detection framework, allowing efficient adaptation to \hoi detection. 
Extensive experiments demonstrate the effectiveness of our design and show that \ours achieves state-of-the-art performance on both the popular \swig~\cite{SWiG-HOI} and \hicodet~\cite{HICO-DET} benchmarks.

\section{Related Work}
\label{sec:related_work}

\subsection{\hoi Detector Structures}
Existing methods decompose \hoi detection into two sub-tasks: object detection and interaction classification.
Based on this division, \hoi detection architectures are commonly grouped into two-stage and one-stage ones.

The two-stage models~\cite{CHOIR,SCG,PD-NET,PMFNet,DRG} typically employ an existing object detector to first localize humans and objects, and then perform human-object pairing and interaction classification in the second stage.
 Various features can support interaction classification, including visual features~\cite{CHOIR}, spatial features~\cite{SCG}, human pose~\cite{PD-NET,PMFNet}, and language features~\cite{DRG}.
The two-stage methods have the advantage of a clear model structure: humans and objects are detected first, allowing the second stage to focus on interaction classification.
Their main disadvantage is inefficiency in enumerating human-object pairs and potential error propagation from inaccurate detections in the first stage.

The one-stage methods perform interactive human-object detection and interaction classification in a single forward pass.
Early one-stage designs represent an interacting human-object pair by an interaction region, e.g., a single point~\cite{PPDM}, a set of points~\cite{GGNET}, and the union box of a human-object pair~\cite{UnionDet}.
Modern designs typically adopt the Detection Transformer (DETR)~\cite{DETR} as the backbone, and pre-train its parameters for the object detection task. 
Thanks to the powerful cross-attention mechanism in DETR, these methods can represent an interaction region more flexibly and incorporate more image-level context.
Moreover, many variants of the \hoi queries have been developed in DETR: some~\cite{QPIC,DOQ} adopt a single query to predict the human, the object, and the interaction category in an \hoi triplet simultaneously, while others~\cite{CDN, GEN-VLKT} employ independent queries for the three elements.

Our approach falls into the one-stage paradigm. Unlike most existing one-stage approaches, our objective is to design a simple and streamlined architecture that is strong for both open-vocabulary and closed-set \hoi detection.

\subsection{Open-Vocabulary \hoi Detection}
The annotation of \hoi triples is time-consuming, which affects the diversity of \hoi categories in the training data.
Therefore, recent research has increasingly focused on \ov \hoi detection, which aims to recognize \hoi triplets that are even unseen during training. 
The two-stage \hoi detection methods~\cite{CMMP,EZ-HOI,HOLa} are usually straightforward to extend to recognize unseen \hoi categories.
They typically adopt an off-the-shelf object detector to perform human and object detection in the first stage, and use a VLM to classify interactions within the detected human-object region in the second stage. 
The one-stage \ov \hoi detection methods are more diverse and can be grouped into three categories. The first category of methods is based on compositional learning~\cite{VCL,SCL}, which encourages models to be generalizable by recombining seen \emph{\textless human-verb-object\textgreater} triplets.
The second category of methods~\cite{RLIPv2,DP-HOI} employs large-scale retraining to enhance the generalization ability of HOI detectors.
The third category of methods resorts to VLMs to obtain robust representations for unseen HOI categories.
Moreover, VLM-based approaches can be divided into two types.
The first type of approaches~\cite{GEN-VLKT,HOICLIP,UniHOI,BC-HOI} retains a conventional HOI detector for interactive human-object localization and leverages a VLM mainly for interaction classification.
Since two separately trained models are adopted, these approaches tend to be more complex in their architectures and struggle to fuse features across the two models.
To alleviate this problem, the second type of approaches~\cite{THID,CMD-SE,SGC-Net,INP-CC} employs a single VLM for both interactive human-object detection and interaction classification.
Although they excel at open-vocabulary interaction classification, their detection performance is often weak due to a lack of fine-grained visual features. 
This is because most VLMs are pre-trained for image-level tasks, and their internal representations often lack the precise, region-specific details required for object localization.

In this paper, we also rely on a single VLM for \ov \hoi detection. Unlike existing approaches, we adopt the latest \dino model~\cite{DINOv3} as the backbone.
We streamline it for \hoi detection and carefully address the representation gaps between modules, achieving the state-of-the-art \ov \hoi detection performance. 

\section{Preliminaries}
\label{sec:preliminaries}
\paragraph{\dino.}
\dino~\cite{DINOv3} leverages self-distillation to learn rich feature representations and employs the Gram anchoring strategy to preserve dense spatial details during large-scale self-supervised training.
It uses a Vision Transformer (ViT) \cite{ViT} architecture with $L$ self-attention layers.
It splits an input image $I \in \mathbb{R}^{H \times W \times 3}$ into non-overlapping patches, projects them into patch tokens, and augments them with positional embeddings.
These image patch tokens, a \cls token $\mathbf{x}_{\text{cls}}$, and a set of register tokens $\mathbf{x}_{\text{reg}}$ \cite{Registers} form the input sequence of ViT.
The VIT’s output (denoted as final $\mathbf{Z}_L$) contains contextualized features for all image tokens.

\paragraph{\dinotxt.}  
\dinotxt~\cite{dinotxt} extends the ViT-L/16 backbone of \dino with a vision head and a text encoder for Locked-image Tuning (LiT)~\cite{LiT}. 
The vision head consists of two self-attention blocks that refine backbone features, jointly processing the class token, register tokens, and patch tokens to enhance inter-token dependencies and project visual features into the shared text embedding space. 
This increases the semantic richness of patch tokens while slightly reducing local details. During LiT, the \dino backbone is frozen and only the text encoder and the vision head are trained. 
Text-image alignment is enforced by aligning both the class token and the mean-pooled patch features with the corresponding text embeddings.

\section{Method}
\label{sec:method}

\subsection{Overview}

\paragraph{Motivations.}

\begin{figure}[t]
    \centering
    \begin{subfigure}[b]{1.0\linewidth}
        \includegraphics[width=\textwidth]{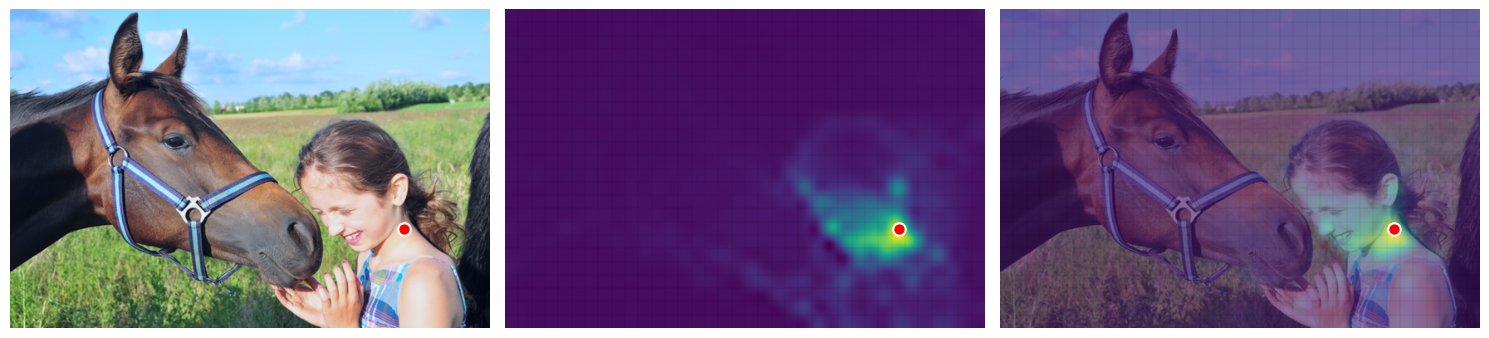}
        \caption{\dino backbone.}
        \label{fig:dinov3_backbone}
    \end{subfigure}    
    \begin{subfigure}[b]{1.0\linewidth}
        \includegraphics[width=\textwidth]{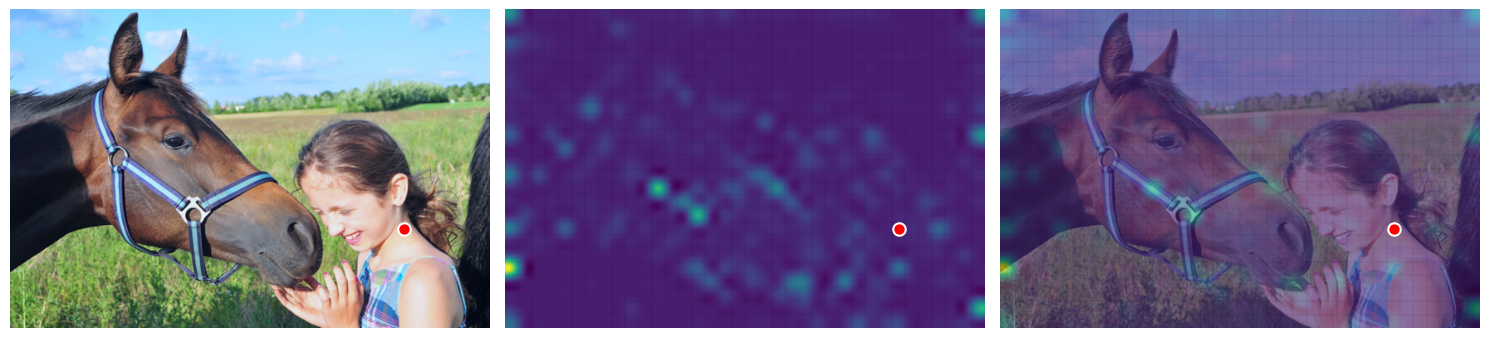}
        \caption{\dinotxt vision head.}
        \label{fig:dino.txt}
    \end{subfigure}
    \caption{Visualization of attention maps from the last self-attention block of (a) \dino backbone and (b) \dinotxt vision head. The left column shows the original image of a person petting a horse, the middle column displays the attention map, and the right column overlays the attention on the original image. The red dot marks the queried patch located on the person. All other image patch tokens are as keys.}
    \label{fig:zero_shot_visualization}
\end{figure}

Our work begins with a fundamental question: can a single, unified model naturally provide the distinct feature types required for both precise localization and broad semantic classification? The attention maps in Fig.~\ref{fig:zero_shot_visualization} provide a compelling affirmative. We observed an explicit functional specialization within the \dinotxt model.
The attention map of the \dino backbone, shown in Fig.~\ref{fig:dinov3_backbone}, is tightly focused, attending to small, specific areas in an image. This focus provides the fine-grained spatial detail essential for instance detection~\cite{ViT-Adapter}. In contrast, the vision head in Fig.~\ref{fig:dino.txt} exhibits holistic attention that aggregates the entire relational context. This allows it to form an ideal foundation for interaction classification~\cite{QPIC}.
This discovery of inherent, complementary roles becomes our architectural principle: we harness the backbone for detailed spatial representation, while the head provides semantic comprehension. This enables us to construct a streamlined one-stage framework.

\begin{figure*}[ht]
  \centering
  \includegraphics[width=\linewidth]{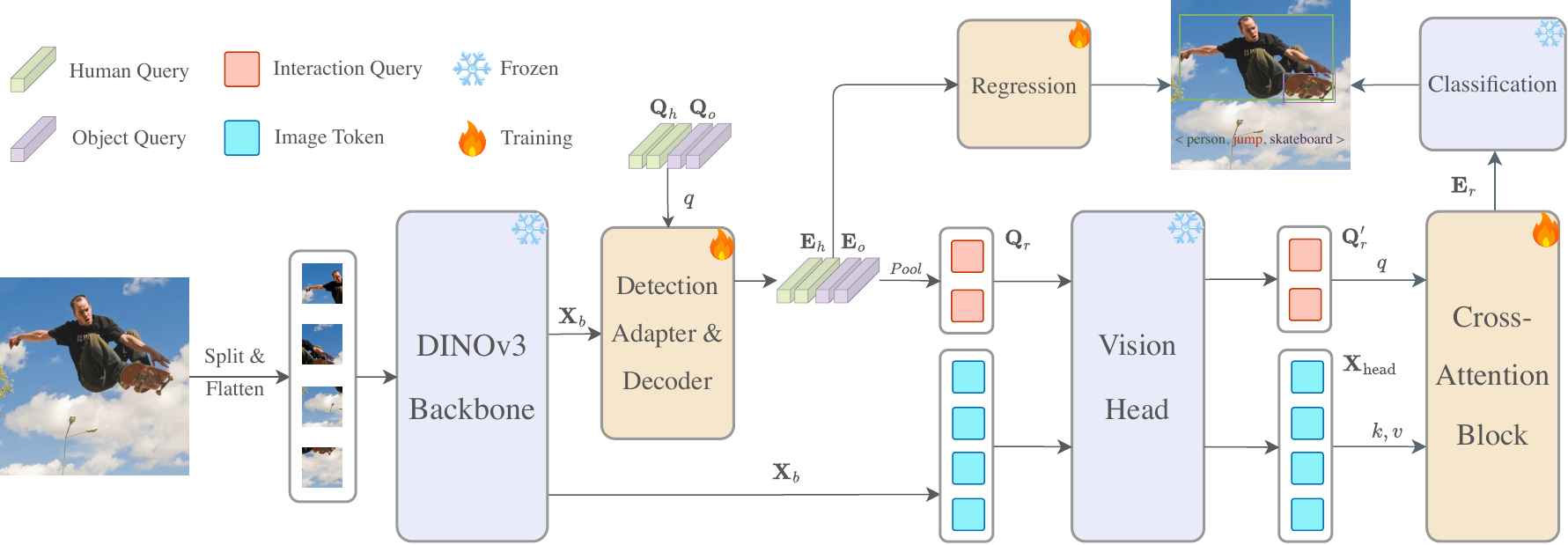}
  \caption{
  Overall architecture of our \ours framework. A frozen \dino ViT encoder (backbone) provides features for two branches. The first branch performs standard instance detection, localizing interactive human-object pairs. The second branch, our core contribution, refines interaction queries in a two-step process. We feed the initial interaction queries $\mathbf{Q}_r$ along with image tokens into the frozen vision head. This yields semantically enriched queries $\mathbf{Q}_r'$ and contextualized image tokens $\mathbf{X}_{\text{head}}$. Subsequently, we employ a single learnable cross-attention block that uses these enriched queries to re-attend to $\mathbf{X}_{\text{head}}$, producing higher-quality embeddings $\mathbf{E}_r$, which are used for open-vocabulary interaction classification.
  }
  \label{fig:framework}
\end{figure*}

\paragraph{Overall Architecture.}
We illustrate the overall architecture of \ours in Fig.~\ref{fig:framework}.
It is a one-stage framework built on the \dino model.
Given an input image $I \in \mathbb{R}^{H \times W \times 3}$, the frozen \dino backbone produces image tokens $\mathbf{X}_b \in \mathbb{R}^{N \times D}$, where $N$ and $D$ denote the token number and the embedding dimension, respectively.
These tokens are used for both interactive human-object instance detection and interaction classification tasks.

For the first task, we adopt the standard detection decoder in \hoi detection works~\cite{GEN-VLKT}. It has one set of learnable human queries $\mathbf{Q}_h \in \mathbb{R}^{N_q \times d}$ and one set of object queries  $\mathbf{Q}_o \in \mathbb{R}^{N_q \times d}$. The obtained decoder embeddings $\mathbf{E}_h$ and $\mathbf{E}_o$ are used to predict the human and object bounding boxes, respectively.

For the second task, we form initial interaction queries $\mathbf{Q}_r \in \mathbb{R}^{N_q \times D}$ by performing element-wise averaging on $\mathbf{E}_h$ and $\mathbf{E}_o$. We reduce the representation gap between $\mathbf{Q}_r$ and the output of the vision head by feeding $\mathbf{Q}_r$ and $\mathbf{X}_b$ together to the frozen vision head, resulting in mutually adapted interaction queries $\mathbf{Q}_r'$ and image tokens $\mathbf{X}_{\text{head}}$. 
Finally, we perform cross-attention between $\mathbf{Q}_r'$ and $\mathbf{X}_{\text{head}}$, and the output decoder embeddings $\mathbf{E}_r$ are used for open-vocabulary interaction classification.

\subsection{Interactive Human-Object Detection}
We first reduce the embedding dimension of $\mathbf{X}_b$ to $d$ using a convolutional layer of $1 \times 1$.
We then add positional encodings $\mathbf{E}_{pos}$ to the image patch tokens. The resulting features are further processed by a detection adapter consisting of $L_E$ self-attention layers. The above process can be formulated as follows:
\begin{equation}
    \label{eq:adapter}
    \mathbf{F} = \mathrm{Adapter}(\mathrm{Conv}(\mathbf{X}_b) + \mathbf{E}_{pos}).
\end{equation}

Then, we adopt a transformer decoder that includes $L_D$ cross-attention layers for interactive human-object detection. It has two independent sets of learnable queries: $\mathbf{Q}_h$ for humans and $\mathbf{Q}_o$ for objects. Both sets of queries adopt $\mathbf{F}$ as the key and value. This yields refined embeddings $\mathbf{E}_h$ and $\mathbf{E}_o$:
\begin{equation}
    \label{eq:det_decoder}
    \mathbf{E}_h, \mathbf{E}_o = \mathrm{Decoder}(\mathbf{Q}_h, \mathbf{Q}_o, \mathbf{F}).
\end{equation}

Finally, we detect the human and object instances as follows:
\begin{equation}
    \hat{b}_h = \mathrm{MLP}_h(\mathbf{E}_h), \quad
    \hat{b}_o = \mathrm{MLP}_o(\mathbf{E}_o),
\end{equation}
where $\hat{b}_h$ and $\hat{b}_o$ denote the regressed human and object bounding boxes, respectively.

\subsection{Interaction Classification}
Although the pre-trained \dino model offers rich features, these are not inherently optimized for classifying human-object interactions. To bridge this gap, we introduce a streamlined two-step process to adapt \dino's features for this task.
We begin this by forming initial interaction queries, $\mathbf{Q}_r$, by projecting the mean of interactive human-object pair embeddings to the dimension of the frozen vision head:
\begin{equation}
    \label{eq:initial_interaction_queries}
    \mathbf{Q}_r = \mathrm{Proj}(\left(\mathbf{E}_h + \mathbf{E}_o\right) / 2).
\end{equation}

\paragraph{Semantic Bootstrapping in the Frozen Vision Head.}
In the first step, we bootstrap the interaction queries, $\mathbf{Q}_r$, using high-level semantic context from the frozen vision head, denoted as $\mathcal{F}_{\text{head}}$. 

The interaction queries $Q_r$ are concatenated with the backbone's image tokens, $\mathbf{X}_{b}$, and passed through the vision head's self-attention layers at no additional training cost:
\begin{equation}
\label{eq:semantic_bootstrapping}
    [\mathbf{Q}_r'; \mathbf{X}_{\text{head}}] = \mathcal{F}_{\text{head}}([\mathbf{Q}_r; \mathbf{X}_{b}]).
\end{equation}
This operation yields two key outputs. The first is a set of semantically enriched interaction queries, $\mathbf{Q}_r'$, aligned with the head's text-semantic space. The second is a set of \textit{query-influenced} image tokens, $\mathbf{X}_{\text{head}}$. These image tokens are now contextually modulated by task-specific interaction queries.

\paragraph{Hierarchical Refinement via a Cross-Attention Block.}
Although bootstrapped queries $\mathbf{Q}_r'$ alone improve classification performance, a streamlined architecture should take advantage of all available information.
The query-influenced image tokens, $\mathbf{X}_{\text{head}}$, represent a valuable contextualized feature source that should not be ignored. 
In the second step, we leverage the contextualized information in $\mathbf{X}_{\text{head}}$. We introduce a lightweight, learnable decoder, $\mathcal{G}_{\text{decoder}}$, that refines the enriched queries $\mathbf{Q}_r'$ by conditioning on image tokens influenced by them:
\begin{equation}
\label{eq:hierarchical_refinement}
    \mathbf{E}_r = \mathcal{G}_{\text{decoder}}(\mathbf{Q}_r', \mathbf{X}_{\text{head}}).
\end{equation}
Composed of a single learnable cross-attention layer and an MLP layer, this decoder distills the most salient cues from contextualized tokens, producing the final specialized decoder embeddings $\mathbf{E}_r$ for open-vocabulary classification.
This hierarchical process—a coarse semantic alignment followed by a focused, learnable refinement—is key to the performance of our method.
The resulting decoder embeddings $\mathbf{E}_r$ are therefore maximally informed and specialized for accurate \ov classification of interactions.

\paragraph{\OV Predictions.}
For the final classification, the refined interaction decoder embeddings $\mathbf{E}_r = \{\mathbf{e}_r^{(i)}\}$ are first mapped to the text embedding space via a linear projection layer. Let the projected embeddings be $\mathbf{e}_r'^{(i)}$. We then compute class probabilities by measuring the cosine similarity between these projected embeddings and the text embeddings $\mathbf{E}_t = \{\mathbf{e}_t^{(j)}\}$, which are pre-computed for all interaction categories using a frozen text encoder. The probability is given by:
\begin{equation}
    \label{eq:open_vocab_classification}
    p_{ij} = \frac{\exp(\tau \cdot \cos(\mathbf{e}_r'^{(i)}, \mathbf{e}_t^{(j)}))}{\sum_{k \in \mathcal{R}} \exp(\tau \cdot \cos(\mathbf{e}_r'^{(i)}, \mathbf{e}_t^{(k)}))},
\end{equation}
where $p_{ij}$ denotes the probability that the $i$-th interaction representation is classified into the $j$-th category, $\tau$ is a learnable temperature, and $\mathcal{R}$ is the set of all interaction categories.
This yields the final \ov interaction classification results.

\section{Experiments}
\label{sec:experiments}

\subsection{Datasets and Metrics}

\paragraph{Datasets.}
We conduct experiments on two widely used benchmarks, \swig\cite{SWiG-HOI} and \hicodet\cite{HICO-DET}.
The \swig dataset provides diverse human-object interactions across 406 action and 1,000 object categories.
Its test set contains about 14,000 images and roughly 5,500 relation categories, among which over 1,000 relations are unseen during training, making it a suitable benchmark for \ov \hoi detection.
The \hicodet dataset consists of 600 relation categories, formed by combining 117 action categories and 80 object categories, where the object categories are defined following \coco\cite{COCO}.
In the \ov setting, we follow \cite{VCL,THID} to remove 120 rare interaction categories from the training set while retaining them in the test set.
\paragraph{Evaluation Metrics.}
We follow the settings of previous work \cite{THID,HICO-DET,GEN-VLKT} and use mean Average Precision (mAP) for the evaluation.
We define a true positive when both human and object bounding boxes have an Intersection over Union (IoU) greater than 0.5 with the ground truth, and the predicted interaction label matches the ground truth.

\subsection{Implementation Details}
\label{subsec:implementation_details}
We adopt the ViT-L/16 variant of \dino as our visual backbone. To facilitate a fair comparison, its parameter count is comparable to that of the CLIP model~\cite{CLIP} with a ViT-L/14 backbone.
The detection adapter consists of $L_E = 2$ self-attention layers, and the detection feature dimension is set to $d = 256$.
The instance decoder is composed of $L_D = 3$ layers, and we use $N_q = 64$ learnable queries for human and object instances. The $\mathcal{G}_{\text{decoder}}$ before the interaction classification is a $1$-layer transformer decoder, and its feature dimension is $D = 1024$.
For training, we follow the settings of previous works: \cite{THID} for the \swig dataset and \cite{GEN-VLKT} for the \hicodet dataset. 
Specifically, \swig\ adopts a contrastive objective in which in-batch negatives are used for training, whereas \hicodet\ is trained to classify interactions across the entire category set.
The model is optimized with AdamW \cite{AdamW} using a learning rate of $1 \times 10^{-4}$.
All experiments are conducted on 8 NVIDIA RTX 4090 GPUs, with a batch size of 32 per GPU for \swig and 2 per GPU for \hicodet.

\subsection{Comparison in the \OV Settings}
We evaluate the performance of our model on both the \swig and \hicodet datasets.
Following the experimental settings in \cite{THID} and \cite{GEN-VLKT}, we conduct comparisons with existing methods from multiple perspectives.

\paragraph{\swig.}
As presented in Tab.~\ref{tab:swig_results}, \ours establishes a new state-of-the-art across all metrics.
We attribute this success not only to a stronger vision backbone but also to the intrinsic design of our model.
Specifically, on the rare and non-rare categories, \ours outperforms MP-HOI-L~\cite{MP-HOI}, the previous leading method in these categories by 6.10\% and 4.86\%, respectively.
The generalization capability of our model is further highlighted by its performance on the unseen category, where it surpasses the second-best method, SGC-Net~\cite{SGC-Net}, by 6.58\%.
This advantage is maintained in the full category, where \ours achieves a 7.47\% improvement over SGC-Net.
To isolate the contribution of our model's architecture, we note that simply equipping more powerful backbones does not yield equivalent performance.
For instance, even when augmented with larger backbones such as Swin-Large~\cite{Swin} and CLIP-ViT-L/14 along with additional pre-training, MP-HOI-L does not achieve commensurate gains.
This result underscores that the streamlined architectural design of \ours is uniquely effective in leveraging the rich features of \dino for \ov \hoi detection.

\begin{table}[ht]
\centering
\footnotesize
\setlength{\tabcolsep}{4pt}
\renewcommand{\arraystretch}{1.1}
\caption{Comparison on the \swig dataset (mAP \%).}
\label{tab:swig_results}
\resizebox{\columnwidth}{!}{\begin{tabular}{lcccc}
\toprule
\textbf{Method} & \textbf{Unseen} & \textbf{Rare} & \textbf{Non-rare} & \textbf{Full} \\
\midrule
\multicolumn{5}{l}{\textit{With object detection pre-training}} \\
\midrule
QPIC~\cite{QPIC} & 6.21 & 10.84 & 16.95 & 11.12 \\
GEN-VLKT~\cite{GEN-VLKT} & - & 10.41 & 20.91 & 10.87 \\
MP-HOI-S~\cite{MP-HOI} & - & 14.78 & 20.28 & 12.61 \\
MP-HOI-L~\cite{MP-HOI} & - & \underline{18.59} & \underline{25.76} & 16.21 \\
\midrule
\multicolumn{5}{l}{\textit{Without object detection pre-training}} \\
\midrule
THID~\cite{THID} & 10.04 & 12.82 & 17.67 & 13.26 \\
CMD-SE~\cite{CMD-SE} & 10.70 & 14.64 & 21.46 & 15.26 \\
SGC-Net~\cite{SGC-Net} & \underline{12.46} & 16.55 & 23.67 & \underline{17.20} \\
INP-CC~\cite{INP-CC} & 11.02 & 16.74 & 22.84 & 16.74 \\
Ours & \textbf{19.04} & \textbf{24.69} & \textbf{30.62} & \textbf{24.67} \\
\bottomrule
\end{tabular}}
\end{table}

\paragraph{HICO-DET, \OV Setting.}
The results of our method on the \hicodet dataset are presented in Tab.~\ref{tab:hicodet_results_open_vocabulary}.
In the \hicodet \ov setting, object labels are still derived from \coco~\cite{COCO}, so methods pre-trained on \coco object detection tend to perform better due to the overlapping label space~\cite{SGC-Net,INP-CC}.
We therefore report two groups in Tab.~\ref{tab:hicodet_results_open_vocabulary} for fair comparison.
Even under this biased condition, \ours achieves a strong performance.
Compared with methods that use object detection pre-training, \ours achieves improvements of 2.16\% and 1.50\% in the seen and full categories, respectively.
While BC-HOI~\cite{BC-HOI} reports higher performance in the unseen category, our method remains overall competitive.
Compared with approaches without object detection pre-training, \ours achieves larger gains of 17.26\%, 14.65\%, and 15.27\% in the unseen, seen, and full categories, respectively.
These results clearly demonstrate the robustness of our method in different datasets and evaluation settings.

\begin{table}[ht]
\centering
\footnotesize
\setlength{\tabcolsep}{3pt}
\renewcommand{\arraystretch}{1.2}
\caption{Comparison on the \hicodet dataset in the \ov setting (mAP \%).}
\label{tab:hicodet_results_open_vocabulary}
\makebox[\columnwidth][c]{
\resizebox{1.0\columnwidth}{!}{\begin{tabular}{llccc}
\toprule
\textbf{Method} & \textbf{Backbone} & \textbf{Unseen} & \textbf{Seen} & \textbf{Full} \\
\midrule
\multicolumn{5}{l}{\textit{With object detection pre-training}} \\
\midrule
GEN-VLKT~\cite{GEN-VLKT} & ResNet50 + CLIP-ViT-B/32 & 21.36 & 32.91 & 30.56 \\
HOICLIP~\cite{HOICLIP} & ResNet50 + CLIP-ViT-B/32 & 25.53 & 34.85 & 32.99 \\
CLIP4HOI~\cite{CLIP4HOI} & ResNet50 + CLIP-ViT-B/16 & 28.47 & 35.48 & 34.08 \\
LOGICHOI~\cite{LOGICHOI} & ResNet50 + CLIP-ViT-B/32 & 25.97 & 34.93 & 33.17 \\
UniHOI~\cite{UniHOI} & ResNet50 + BLIP-2-ViT-G/14 & 28.68 & 33.16 & 32.27 \\
BCOM~\cite{BCOM} & ResNet50 + CLIP-ViT-L/14 & 28.52 & 35.04 & 33.74 \\
CMMP~\cite{CMMP} & ResNet50 + CLIP-ViT-L/14 & 35.98 & 37.42 & 37.13 \\
EZ-HOI~\cite{EZ-HOI} & ResNet50 + CLIP-ViT-L/14 & 34.24 & 37.35 & 36.73 \\
HOLa~\cite{HOLa} & ResNet50 + CLIP-ViT-B/16 & 30.61 & 35.08 & 34.19 \\
BC-HOI~\cite{BC-HOI} & ResNet50 + BLIP-2-ViT-G/14 & \textbf{42.31} & \underline{40.67} & \textbf{40.99} \\
VRDiff~\cite{VRDiff} & ResNet50 + CLIP-ViT-L/14 & \underline{38.92} & \textbf{40.83} & \underline{40.45} \\
\midrule
\multicolumn{5}{l}{\textit{Without object detection pre-training}} \\
\midrule
THID~\cite{THID} & CLIP-ViT-B/16 & 15.53 & 24.32 & 22.96 \\
CMD-SE~\cite{CMD-SE} & CLIP-ViT-B/16 & 16.70 & 23.95 & 22.35 \\
SGC-Net~\cite{SGC-Net} & CLIP-ViT-B/16 & \underline{23.27} & \underline{28.34} & \underline{27.22} \\
INP-CC~\cite{INP-CC} & CLIP-ViT-B/16 & 17.38 & 24.74 & 23.13 \\
Ours & DINOv3-ViT-L/16 & \textbf{40.53} & \textbf{42.99} & \textbf{42.49} \\
\bottomrule
\end{tabular}}
}
\end{table}

\subsection{Comparison in the Closed Setting}
We further evaluate our method in the closed setting of the \hicodet dataset following \cite{GEN-VLKT}, where all 600 interaction categories are present during training.
We compare our results with two-stage and one-stage \hoi detection methods, as summarized in Tab.~\ref{tab:hicodet_results_closed}.
\ours outperforms all other methods in this setting.
Specifically, \ours surpasses the previous state-of-the-art BC-HOI~\cite{BC-HOI}, with significant gains of +2.04\% on the full set, +1.95\% on rare, and +2.07\% on non-rare categories.
Notably, this robust performance in the closed setting is achieved without relying on object detection pre-training from datasets like COCO, demonstrating the powerful convergence capabilities and inherent strength of our framework.

\begin{table}[ht]
\centering
\footnotesize
\setlength{\tabcolsep}{4pt}
\renewcommand{\arraystretch}{1.3}
\caption{Comparison on the \hicodet dataset in the closed setting (mAP \%).}
\label{tab:hicodet_results_closed}
\makebox[\columnwidth][c]{
\resizebox{1.0\columnwidth}{!}{\begin{tabular}{llccc}
\toprule
\textbf{Method} & \textbf{Backbone} & \textbf{Rare} & \textbf{Non-rare} & \textbf{Full} \\
\midrule
\multicolumn{5}{l}{\textit{Two-stage methods}} \\
\midrule
UPT~\cite{UPT} & ResNet50 & 25.94 & 33.36 & 31.66 \\
CLIP4HOI~\cite{CLIP4HOI} & ResNet50 + CLIP-ViT-B/16 & 33.95 & 35.74 & 35.33 \\
CMMP~\cite{CMMP} & ResNet50 + CLIP-ViT-L/14 & 37.75 & 38.25 & 38.14 \\
BCOM~\cite{BCOM} & ResNet50 + CLIP-ViT-L/14 & 39.90 & 39.17 & 39.34 \\
EZ-HOI~\cite{EZ-HOI} & ResNet50 + CLIP-ViT-L/14 & 37.70 & 38.89 & 38.61 \\
HOLa~\cite{HOLa} & ResNet50 + CLIP-ViT-L/14 & 38.66 & 39.18 & 39.05 \\
VRDiff~\cite{VRDiff} & ResNet50 + CLIP-ViT-L/14 & 41.69 & 41.31 & 41.40 \\
\midrule
\multicolumn{5}{l}{\textit{One-stage methods}} \\
\midrule
QPIC~\cite{QPIC} & ResNet50 & 21.85 & 31.23 & 29.07 \\
CDN~\cite{CDN} & ResNet50 & 27.39 & 32.64 & 31.44 \\
GEN-VLKT~\cite{GEN-VLKT} & ResNet50 + CLIP-ViT-B/32 & 29.25 & 35.10 & 33.75 \\
DOQ~\cite{DOQ} & ResNet50 + CLIP-ViT-B/16 & 29.19 & 34.50 & 33.28 \\
LOGICHOI~\cite{LOGICHOI} & ResNet50 + CLIP-ViT-B/32 & 32.03 & 36.22 & 35.47 \\
HOICLIP~\cite{HOICLIP} & ResNet50 + CLIP-ViT-B/32 & 31.12 & 35.74 & 34.69 \\
FGAHOI~\cite{FGAHOI} & Swin-Large & 30.71 & 39.11 & 37.18 \\
DP-HOI~\cite{DP-HOI} & ResNet50 + CLIP-ViT-B/32 & 34.36 & 37.22 & 36.56 \\
UniHOI~\cite{UniHOI} & ResNet50 + BLIP-2-ViT-G/14 & 39.91 & 40.11 & 40.06 \\
BC-HOI~\cite{BC-HOI} & ResNet50 + BLIP-2-ViT-G/14 & \underline{45.76} & \underline{42.18} & \underline{43.01} \\
Ours & DINOv3-ViT-L/16 & \textbf{47.71} & \textbf{44.25} & \textbf{45.05} \\
\bottomrule
\end{tabular}}
}
\end{table}

\subsection{Ablation Studies}
\begin{figure*}[ht]
  \centering
  \includegraphics[width=\linewidth]{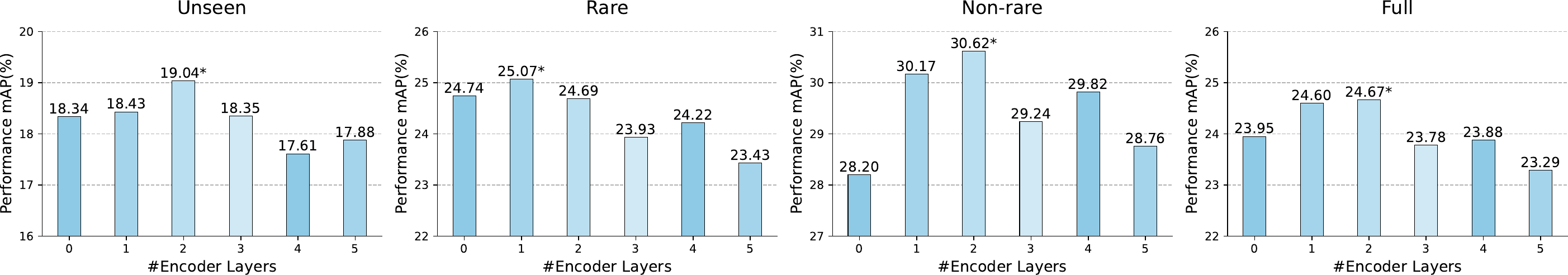}
  \caption{Ablation studies on the number of encoder layers in the detection adapter on the \swig dataset (mAP \%).}
  \label{fig:swig_ablation_on_encoder_layers}
\end{figure*}

We conduct comprehensive ablation studies on the \swig dataset to validate the core design of our framework.
Our analysis is threefold.
First, we perform an additive analysis to quantify the contribution of each key architectural component, with results presented in Tab.~\ref{tab:swig_ablation_on_architecture}.
Second, we compare our final model against several plausible design variants to justify our specific architectural choices, as summarized in Tab.~\ref{tab:swig_ablation_on_method}.
Finally, we analyze the impact of varying the number of encoder layers in our detection adapter, as illustrated in Fig.~\ref{fig:swig_ablation_on_encoder_layers}.

\paragraph{Architecture Design.}
Tab.~\ref{tab:swig_ablation_on_architecture} presents ablation studies on the key architectural components of our method.
The table begins with a strong baseline model, whose architecture is illustrated in the supplementary material.
To construct this baseline, we replace our proposed interaction classification module with a more conventional design, such as that in HOICLIP~\cite{HOICLIP}.
Specifically, a $3$-layer transformer decoder is used to perform cross-attention between interaction queries and the semantic features of the frozen vision head.
We call this fusion strategy late-fusion hereafter. 
All other parameters remain identical to those of our final model.
In particular, this baseline already achieves a high performance of 16.55\%, 21.66\%, 27.75\%, and 21.82\% on the unseen, rare, non-rare, and full categories of the \swig dataset, respectively, which we attribute to the powerful representations provided by the \dino backbone.

Next, we replace this late-fusion decoder with our \textbf{Semantic Bootstrapping}. In this step, the interaction queries are processed alongside the image tokens within the frozen vision head. 
This allows the queries to benefit from the pre-trained head parameters directly and to interact fully with the image tokens via the head's self-attention blocks.
This single change yields substantial gains of +1.54\%, +1.61\%, +1.08\%, and +1.46\% across the unseen, rare, non-rare, and full categories.
Finally, we introduce \textbf{Hierarchical Refinement}, which re-utilizes the query-influenced image tokens produced by the previous stage. The interaction queries re-attend to these contextualized tokens, forming our complete framework \ours. This step further improves performance by +0.95\%, +1.42\%, +1.79\%, and +1.39\%.

The analysis reveals a clear division of benefits.
Semantic Bootstrapping shares the rich semantic space of the frozen head, significantly boosting generalization on unseen and rare categories.
Hierarchical Refinement, on the other hand, leverages image tokens cued by the \hoi detection task, yielding larger gains across rare and non-rare categories.
In total, \ours achieves cumulative improvements of +2.49\%, +3.03\%, +2.87\%, and +2.85\% over the strong baseline. This clearly demonstrates the effectiveness of our design and its ability to successfully adapt the powerful \dino model for the \ov \hoi detection task.

\begin{table}[ht]
\centering
\footnotesize
\setlength{\tabcolsep}{4pt}
\renewcommand{\arraystretch}{1.1}
\caption{Ablation study of our model's architectural components on the \swig dataset (mAP \%).}
\label{tab:swig_ablation_on_architecture}
\makebox[\columnwidth][c]{
\resizebox{1.0\columnwidth}{!}{\begin{tabular}{lcccc}
\toprule
\textbf{Configuration} & \textbf{Unseen} & \textbf{Rare} & \textbf{Non-rare} & \textbf{Full} \\
\midrule
Baseline & 16.55 & 21.66 & 27.75 & 21.82 \\
+ Semantic Bootstrapping & 18.09 & 23.27 & 28.83 & 23.28 \\
+ Hierarchical Refinement & \textbf{19.04} & \textbf{24.69} & \textbf{30.62} & \textbf{24.67} \\
\bottomrule
\end{tabular}}
}
\end{table}

\paragraph{Variants of SL-HOI.}
Our method can be conceptually understood as a form of multi-scale feature fusion, combining features from before and after the vision head.
However, our approach makes two critical distinctions from conventional multi-scale designs:
1) Rather than only fusion with the final output features, we leverage the head's internal, pre-trained computational forward pathway.
2) Task-specific interaction queries contextually modulate the image tokens processed by the vision head.
To validate the importance of these two design choices, we introduce several variants in our ablation study, with results summarized in Tab.~\ref{tab:swig_ablation_on_method}.

To investigate the first point, we compare our method against two alternatives that use a learnable decoder for fusion rather than our Semantic Bootstrapping.
The first variant, labeled ``Late Fusion (Head only)'', serves as our baseline model, in which a decoder performs cross-attention solely over the head's output tokens.
The second ``Late Fusion (Multi-Scale)'' extends this by attending to both the backbone and the head output tokens. 
As shown in Tab.~\ref{tab:swig_ablation_on_method}, both of these learnable fusion strategies are suboptimal. A single standalone decoder alone struggles to match the performance of our approach.
In contrast, our method effectively transfers the head's generalization capabilities by processing queries directly within its frozen, pre-trained self-attention blocks.

To address the second point, we return to our whole model and introduce a modification labeled ``Ours w/ Attention Mask''.
In this variant, we mask the attention mechanism during Semantic Bootstrapping to prevent the image tokens from being influenced by the interaction queries.
These ``pure'' image tokens, now lacking task-specific cues, lead to a drop in performance across all metrics.
This result is significant: it demonstrates that the interaction queries function not only as information receivers but also as information givers, dynamically refining the image representations for the downstream \hoi detection task.

\begin{table}[ht]
\centering
\footnotesize
\setlength{\tabcolsep}{4pt}
\renewcommand{\arraystretch}{1.1}
\caption{Ablation study of variants of our proposed method on the \swig dataset (mAP \%).}
\label{tab:swig_ablation_on_method}
\makebox[\columnwidth][c]{
\resizebox{1.0\columnwidth}{!}{\begin{tabular}{lcccc}
\toprule
\textbf{Configuration} & \textbf{Unseen} & \textbf{Rare} & \textbf{Non-rare} & \textbf{Full} \\
\midrule
Late Fusion (Head only) & 16.55 & 21.66 & 27.75 & 21.82 \\
Late Fusion (Multi-Scale) & 15.73 & 21.63 & 28.49 & 21.77 \\
Semantic Bootstrapping & 18.09 & 23.27 & 28.83 & 23.28 \\
Ours w/ Attention Mask & 17.28 & 24.64 & 29.81 & 24.01 \\
\textbf{Ours} & \textbf{19.04} & \textbf{24.69} & \textbf{30.62} & \textbf{24.67} \\
\bottomrule
\end{tabular}}
}
\end{table}

\paragraph{Number of encoder layers.}
Fig.~\ref{fig:swig_ablation_on_encoder_layers} presents the ablation studies on the number of encoder layers in our detection adapter.
Adapting pre-trained \dino features for downstream tasks, particularly for dense prediction, requires a delicate balance. Since \dino's representations are learned via self-supervision, its parameters are kept frozen to preserve their quality.
However, a frozen backbone presents a challenge for DETR-based~\cite{DETR} architectures, which benefit from end-to-end optimization. This makes the number of encoder layers in the detection adapter a critical hyperparameter.
Using too many layers risks corrupting the rich \dino features, while using too few may not adequately adapt them to the demands of the HOI detection task.

As illustrated in Fig.~\ref{fig:swig_ablation_on_encoder_layers}, simply increasing the number of encoder layers does not lead to better performance.
This finding contrasts with the original DETR paper's conclusion~\cite{DETR}, which found that deeper encoders generally yield monotonic performance gains.
We observe that setting the number of encoder layers to $2$ achieves the best trade-off, yielding the highest performance on the full category.
Therefore, we use two encoder layers in all our experiments.

\subsection{Qualitative Analysis}
We also perform qualitative experiments to evaluate SL-HOI, focusing on analyzing the attention maps for our two-step interaction classification.

\begin{figure}[t]
    \centering
    \captionsetup[subfigure]{skip=-10pt}
    \begin{subfigure}[b]{0.325\columnwidth}
        \includegraphics[width=\textwidth]{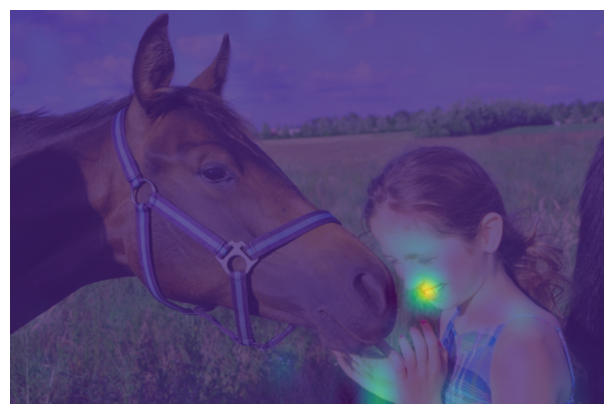}
        \label{fig:sa0}
        \caption{1st Self-Attention}
    \end{subfigure}
    \hfill
    \begin{subfigure}[b]{0.325\columnwidth}
        \includegraphics[width=\textwidth]{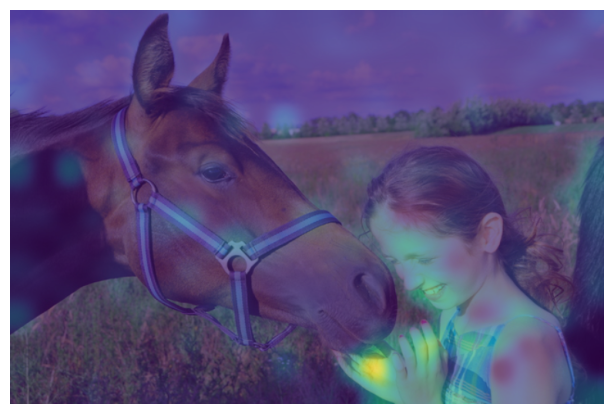}
        \label{fig:sa1}
        \caption{2nd Self-Attention}        
    \end{subfigure}
    \hfill
    \begin{subfigure}[b]{0.325\columnwidth}
        \includegraphics[width=\textwidth]{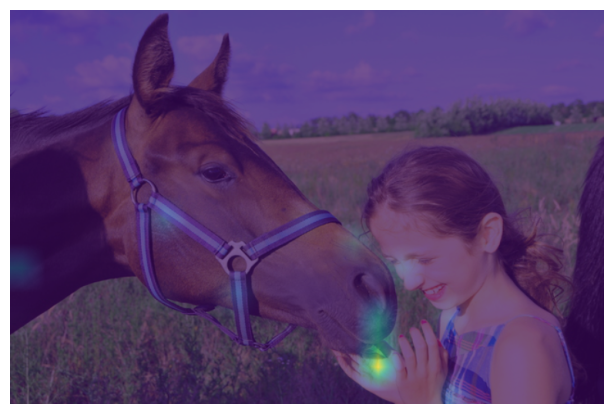}
        \label{fig:ca0}
        \caption{Cross Attention}
    \end{subfigure}
    \captionsetup{skip=4pt}
    \caption{Visualization of attention maps across the interaction classification stage. The left two are in the self-attention blocks of the frozen head during Semantic Bootstrapping, and the right one is from the cross-attention block in Hierarchical Refinement, illustrating a Local-Global-Local interaction reasoning process.}
    \label{fig:classification_attn}
\end{figure}

As shown in Fig.~\ref{fig:classification_attn}, the attention map during the semantic bootstrapping stage characteristically covers a broad area. This behavior stems mainly from the frozen vision head, which was pre-trained to align image patch tokens with textual captions. This objective encourages broader information exchange among tokens, leading to greater attention to capture rich contextual information.
As the model transitions to the hierarchical refinement stage, the nature of the attention shifts. We identify two factors that drive this adaptation:
(1) The preceding semantic bootstrapping stage aligns interaction queries and image tokens within a shared semantic space, which helps guide the attention toward potential interaction regions.
(2) The now-unfrozen, learnable cross-attention block allows the mechanism to specialize for the \hoi detection objective, refining its focus from the broader context.
This two-stage process yields final attention maps that balance contextual understanding with a focus on salient interaction cues, a distinct characteristic of our model's decoding process.

\section{Conclusion and Limitations}
\label{sec:conclusion}
In this paper, we present \ours, a streamlined one-stage framework for \ov \hoi detection built upon the \dino model. We leverage the complementary strengths of \dino's backbone and vision head to effectively address both interactive human-object detection and open-vocabulary interaction classification tasks. Our design includes a novel two-step interaction classification process that bridges representation gaps and enhances feature utilization. Extensive experiments on two popular benchmarks demonstrate that \ours achieves state-of-the-art performance in open-vocabulary \hoi detection while maintaining a simple architecture with few trainable parameters. Our work has certain limitations. For example, using the ViT backbone in the DINOv3 model may incur higher computational costs than traditional CNN-based HOI detectors.

\vspace{0.2em}
\paragraph{Broader Impacts.} 
By advancing \hoi detection, our work can benefit many fields, such as robotics and assistive technologies. To the best of our knowledge, our method has no obvious negative social impacts.

\vspace{0.2em}
\paragraph{Acknowledgement.} 
This work was supported by the National Natural Science Foundation of China under Grant 62476099 and 62076101, Guangdong Basic and Applied Basic Research Foundation under Grant 2024B1515020082 and 2023A1515010007, the Guangdong Provincial Key Laboratory of Human Digital Twin under Grant 2022B1212010004, the TCL Young Scholars Program.

\newpage
{
    \small
    \bibliographystyle{ieeenat_fullname}

\begin{thebibliography}{52}
\providecommand{\natexlab}[1]{#1}
\providecommand{\url}[1]{\texttt{#1}}
\expandafter\ifx\csname urlstyle\endcsname\relax
  \providecommand{\doi}[1]{doi: #1}\else
  \providecommand{\doi}{doi: \begingroup \urlstyle{rm}\Url}\fi

\bibitem[Cao et~al.(2025)Cao, Tang, Zhang, Zheng, Liang, Wei, and Zhao]{VRDiff}
Ping Cao, Yepeng Tang, Chunjie Zhang, Xiaolong Zheng, Chao Liang, Yunchao Wei, and Yao Zhao.
\newblock Visual relation diffusion for human-object interaction detection.
\newblock In \emph{ICCV}, 2025.

\bibitem[Cao et~al.(2023)Cao, Tang, Su, Chen, You, Lu, and Xu]{UniHOI}
Yichao Cao, Qingfei Tang, Xiu Su, Song Chen, Shan You, Xiaobo Lu, and Chang Xu.
\newblock Detecting any human-object interaction relationship: Universal {HOI} detector with spatial prompt learning on foundation models.
\newblock In \emph{NeurIPS}, 2023.

\bibitem[Carion et~al.(2020)Carion, Massa, Synnaeve, Usunier, Kirillov, and Zagoruyko]{DETR}
Nicolas Carion, Francisco Massa, Gabriel Synnaeve, Nicolas Usunier, Alexander Kirillov, and Sergey Zagoruyko.
\newblock End-to-end object detection with transformers.
\newblock In \emph{ECCV}, 2020.

\bibitem[Chao et~al.(2018)Chao, Liu, Liu, Zeng, and Deng]{HICO-DET}
Yu{-}Wei Chao, Yunfan Liu, Xieyang Liu, Huayi Zeng, and Jia Deng.
\newblock Learning to detect human-object interactions.
\newblock In \emph{WACV}, 2018.

\bibitem[Chen et~al.(2023)Chen, Duan, Wang, He, Lu, Dai, and Qiao]{ViT-Adapter}
Zhe Chen, Yuchen Duan, Wenhai Wang, Junjun He, Tong Lu, Jifeng Dai, and Yu Qiao.
\newblock Vision transformer adapter for dense predictions.
\newblock In \emph{ICLR}, 2023.

\bibitem[Darcet et~al.(2024)Darcet, Oquab, Mairal, and Bojanowski]{Registers}
Timoth{\'{e}}e Darcet, Maxime Oquab, Julien Mairal, and Piotr Bojanowski.
\newblock Vision transformers need registers.
\newblock In \emph{ICLR}, 2024.

\bibitem[Dosovitskiy et~al.(2021)Dosovitskiy, Beyer, Kolesnikov, Weissenborn, Zhai, Unterthiner, Dehghani, Minderer, Heigold, Gelly, Uszkoreit, and Houlsby]{ViT}
Alexey Dosovitskiy, Lucas Beyer, Alexander Kolesnikov, Dirk Weissenborn, Xiaohua Zhai, Thomas Unterthiner, Mostafa Dehghani, Matthias Minderer, Georg Heigold, Sylvain Gelly, Jakob Uszkoreit, and Neil Houlsby.
\newblock An image is worth 16x16 words: Transformers for image recognition at scale.
\newblock In \emph{ICLR}, 2021.

\bibitem[Gao et~al.(2020)Gao, Xu, Zou, and Huang]{DRG}
Chen Gao, Jiarui Xu, Yuliang Zou, and Jia{-}Bin Huang.
\newblock {DRG:} dual relation graph for human-object interaction detection.
\newblock In \emph{ECCV}, 2020.

\bibitem[Gupta and Malik(2015)]{VCOCO}
Saurabh Gupta and Jitendra Malik.
\newblock Visual semantic role labeling.
\newblock \emph{arXiv}, abs/1505.04474, 2015.

\bibitem[Hou et~al.(2020)Hou, Peng, Qiao, and Tao]{VCL}
Zhi Hou, Xiaojiang Peng, Yu Qiao, and Dacheng Tao.
\newblock Visual compositional learning for human-object interaction detection.
\newblock In \emph{ECCV}, 2020.

\bibitem[Hou et~al.(2022)Hou, Yu, and Tao]{SCL}
Zhi Hou, Baosheng Yu, and Dacheng Tao.
\newblock Discovering human-object interaction concepts via self-compositional learning.
\newblock In \emph{ECCV}, 2022.

\bibitem[Hu et~al.(2022)Hu, Shen, Wallis, Allen{-}Zhu, Li, Wang, Wang, and Chen]{LoRA}
Edward~J. Hu, Yelong Shen, Phillip Wallis, Zeyuan Allen{-}Zhu, Yuanzhi Li, Shean Wang, Lu Wang, and Weizhu Chen.
\newblock Lora: Low-rank adaptation of large language models.
\newblock In \emph{ICLR}, 2022.

\bibitem[Hu et~al.(2025)Hu, Ding, Sun, Huang, and Xu]{BC-HOI}
Yupeng Hu, Changxing Ding, Chang Sun, Shaoli Huang, and Xiangmin Xu.
\newblock Bilateral collaboration with large vision-language models for open vocabulary human-object interaction detection.
\newblock In \emph{ICCV}, 2025.

\bibitem[Jose et~al.(2025)Jose, Moutakanni, Kang, Baldassarre, Darcet, Xu, Li, Szafraniec, Ramamonjisoa, Oquab, Sim{\'{e}}oni, Vo, Labatut, and Bojanowski]{dinotxt}
Cijo Jose, Th{\'{e}}o Moutakanni, Dahyun Kang, Federico Baldassarre, Timoth{\'{e}}e Darcet, Hu Xu, Daniel Li, Marc Szafraniec, Micha{\"{e}}l Ramamonjisoa, Maxime Oquab, Oriane Sim{\'{e}}oni, Huy~V. Vo, Patrick Labatut, and Piotr Bojanowski.
\newblock Dinov2 meets text: {A} unified framework for image- and pixel-level vision-language alignment.
\newblock In \emph{CVPR}, 2025.

\bibitem[Kim et~al.(2020)Kim, Choi, Kang, and Kim]{UnionDet}
Bumsoo Kim, Taeho Choi, Jaewoo Kang, and Hyunwoo~J. Kim.
\newblock Uniondet: Union-level detector towards real-time human-object interaction detection.
\newblock In \emph{ECCV}, 2020.

\bibitem[Lei et~al.(2024{\natexlab{a}})Lei, Wang, and Tan]{EZ-HOI}
Qinqian Lei, Bo Wang, and Robby~T. Tan.
\newblock {EZ-HOI:} {VLM} adaptation via guided prompt learning for zero-shot {HOI} detection.
\newblock In \emph{NeurIPS}, 2024{\natexlab{a}}.

\bibitem[Lei et~al.(2025{\natexlab{a}})Lei, Wang, and Tan]{HOLa}
Qinqian Lei, Bo Wang, and Robby~T. Tan.
\newblock Hola: Zero-shot hoi detection with low-rank decomposed vlm feature adaptation.
\newblock In \emph{ICCV}, 2025{\natexlab{a}}.

\bibitem[Lei et~al.(2024{\natexlab{b}})Lei, Yin, and Liu]{CMD-SE}
Ting Lei, Shaofeng Yin, and Yang Liu.
\newblock Exploring the potential of large foundation models for open-vocabulary {HOI} detection.
\newblock In \emph{CVPR}, 2024{\natexlab{b}}.

\bibitem[Lei et~al.(2024{\natexlab{c}})Lei, Yin, Peng, and Liu]{CMMP}
Ting Lei, Shaofeng Yin, Yuxin Peng, and Yang Liu.
\newblock Exploring conditional multi-modal prompts for zero-shot {HOI} detection.
\newblock In \emph{ECCV}, 2024{\natexlab{c}}.

\bibitem[Lei et~al.(2025{\natexlab{b}})Lei, Yin, Chen, Peng, and Liu]{INP-CC}
Ting Lei, Shaofeng Yin, Qingchao Chen, Yuxin Peng, and Yang Liu.
\newblock Open-vocabulary hoi detection with interaction-aware prompt and concept calibration.
\newblock In \emph{ICCV}, 2025{\natexlab{b}}.

\bibitem[Li et~al.(2023)Li, Wei, Wang, and Yang]{LOGICHOI}
Liulei Li, Jianan Wei, Wenguan Wang, and Yi Yang.
\newblock Neural-logic human-object interaction detection.
\newblock In \emph{NeurIPS}, 2023.

\bibitem[Li et~al.(2024{\natexlab{a}})Li, Zhang, Lin, Chen, and He]{PGSG}
Rongjie Li, Songyang Zhang, Dahua Lin, Kai Chen, and Xuming He.
\newblock From pixels to graphs: Open-vocabulary scene graph generation with vision-language models.
\newblock In \emph{CVPR}, 2024{\natexlab{a}}.

\bibitem[Li et~al.(2024{\natexlab{b}})Li, Li, Ding, and Xu]{DP-HOI}
Zhuolong Li, Xingao Li, Changxing Ding, and Xiangmin Xu.
\newblock Disentangled pre-training for human-object interaction detection.
\newblock In \emph{CVPR}, 2024{\natexlab{b}}.

\bibitem[Liao et~al.(2020)Liao, Liu, Wang, Chen, Qian, and Feng]{PPDM}
Yue Liao, Si Liu, Fei Wang, Yanjie Chen, Chen Qian, and Jiashi Feng.
\newblock {PPDM:} parallel point detection and matching for real-time human-object interaction detection.
\newblock In \emph{CVPR}, 2020.

\bibitem[Liao et~al.(2022)Liao, Zhang, Lu, Wang, Li, and Liu]{GEN-VLKT}
Yue Liao, Aixi Zhang, Miao Lu, Yongliang Wang, Xiaobo Li, and Si Liu.
\newblock {GEN-VLKT:} simplify association and enhance interaction understanding for {HOI} detection.
\newblock In \emph{CVPR}, 2022.

\bibitem[Lin et~al.(2014)Lin, Maire, Belongie, Hays, Perona, Ramanan, Doll{\'{a}}r, and Zitnick]{COCO}
Tsung{-}Yi Lin, Michael Maire, Serge~J. Belongie, James Hays, Pietro Perona, Deva Ramanan, Piotr Doll{\'{a}}r, and C.~Lawrence Zitnick.
\newblock Microsoft {COCO:} common objects in context.
\newblock In \emph{ECCV}, 2014.

\bibitem[Lin et~al.(2025)Lin, Shi, Yang, Tang, and Zhou]{SGC-Net}
Xin Lin, Chong Shi, Zuopeng Yang, Haojin Tang, and Zhili Zhou.
\newblock Sgc-net: Stratified granular comparison network for open-vocabulary {HOI} detection.
\newblock In \emph{CVPR}, 2025.

\bibitem[Liu et~al.(2021)Liu, Lin, Cao, Hu, Wei, Zhang, Lin, and Guo]{Swin}
Ze Liu, Yutong Lin, Yue Cao, Han Hu, Yixuan Wei, Zheng Zhang, Stephen Lin, and Baining Guo.
\newblock Swin transformer: Hierarchical vision transformer using shifted windows.
\newblock In \emph{ICCV}, 2021.

\bibitem[Loshchilov and Hutter(2019)]{AdamW}
Ilya Loshchilov and Frank Hutter.
\newblock Decoupled weight decay regularization.
\newblock In \emph{ICLR}, 2019.

\bibitem[Ma et~al.(2024)Ma, Wang, Wang, and Wei]{FGAHOI}
Shuailei Ma, Yuefeng Wang, Shanze Wang, and Ying Wei.
\newblock {FGAHOI:} fine-grained anchors for human-object interaction detection.
\newblock \emph{PAMI}, 2024.

\bibitem[Mao et~al.(2023)Mao, Deng, Zhou, Li, Fang, and Li]{CLIP4HOI}
Yunyao Mao, Jiajun Deng, Wengang Zhou, Li Li, Yao Fang, and Houqiang Li.
\newblock {CLIP4HOI:} towards adapting {CLIP} for practical zero-shot {HOI} detection.
\newblock In \emph{NeurIPS}, 2023.

\bibitem[Mascaro et~al.(2023)Mascaro, Sliwowski, and Lee]{ROBOTICS}
Esteve~Valls Mascaro, Daniel Sliwowski, and Dongheui Lee.
\newblock {HOI4ABOT:} human-object interaction anticipation for human intention reading collaborative robots.
\newblock In \emph{CoRL}, 2023.

\bibitem[Ning et~al.(2023)Ning, Qiu, Liu, and He]{HOICLIP}
Shan Ning, Longtian Qiu, Yongfei Liu, and Xuming He.
\newblock {HOICLIP:} efficient knowledge transfer for {HOI} detection with vision-language models.
\newblock In \emph{CVPR}, 2023.

\bibitem[Qu et~al.(2022)Qu, Ding, Li, Zhong, and Tao]{DOQ}
Xian Qu, Changxing Ding, Xingao Li, Xubin Zhong, and Dacheng Tao.
\newblock Distillation using oracle queries for transformer-based human-object interaction detection.
\newblock In \emph{CVPR}, 2022.

\bibitem[Radford et~al.(2021)Radford, Kim, Hallacy, Ramesh, Goh, Agarwal, Sastry, Askell, Mishkin, Clark, Krueger, and Sutskever]{CLIP}
Alec Radford, Jong~Wook Kim, Chris Hallacy, Aditya Ramesh, Gabriel Goh, Sandhini Agarwal, Girish Sastry, Amanda Askell, Pamela Mishkin, Jack Clark, Gretchen Krueger, and Ilya Sutskever.
\newblock Learning transferable visual models from natural language supervision.
\newblock In \emph{ICML}, 2021.

\bibitem[Russakovsky et~al.(2015)Russakovsky, Deng, Su, Krause, Satheesh, Ma, Huang, Karpathy, Khosla, Bernstein, Berg, and Fei{-}Fei]{ImageNet}
Olga Russakovsky, Jia Deng, Hao Su, Jonathan Krause, Sanjeev Satheesh, Sean Ma, Zhiheng Huang, Andrej Karpathy, Aditya Khosla, Michael~S. Bernstein, Alexander~C. Berg, and Li Fei{-}Fei.
\newblock Imagenet large scale visual recognition challenge.
\newblock \emph{IJCV}, 2015.

\bibitem[Siméoni et~al.(2025)Siméoni, Vo, Seitzer, Baldassarre, Oquab, Jose, Khalidov, Szafraniec, Yi, Ramamonjisoa, Massa, Haziza, Wehrstedt, Wang, Darcet, Moutakanni, Sentana, Roberts, Vedaldi, Tolan, Brandt, Couprie, Mairal, Jégou, Labatut, and Bojanowski]{DINOv3}
Oriane Siméoni, Huy~V. Vo, Maximilian Seitzer, Federico Baldassarre, Maxime Oquab, Cijo Jose, Vasil Khalidov, Marc Szafraniec, Seungeun Yi, Michaël Ramamonjisoa, Francisco Massa, Daniel Haziza, Luca Wehrstedt, Jianyuan Wang, Timothée Darcet, Théo Moutakanni, Leonel Sentana, Claire Roberts, Andrea Vedaldi, Jamie Tolan, John Brandt, Camille Couprie, Julien Mairal, Hervé Jégou, Patrick Labatut, and Piotr Bojanowski.
\newblock Dinov3.
\newblock \emph{arXiv}, abs/2508.10104, 2025.

\bibitem[Tamura et~al.(2021)Tamura, Ohashi, and Yoshinaga]{QPIC}
Masato Tamura, Hiroki Ohashi, and Tomoaki Yoshinaga.
\newblock {QPIC:} query-based pairwise human-object interaction detection with image-wide contextual information.
\newblock In \emph{CVPR}, 2021.

\bibitem[Wan et~al.(2019)Wan, Zhou, Liu, Li, and He]{PMFNet}
Bo Wan, Desen Zhou, Yongfei Liu, Rongjie Li, and Xuming He.
\newblock Pose-aware multi-level feature network for human object interaction detection.
\newblock In \emph{ICCV}, 2019.

\bibitem[Wang et~al.(2024)Wang, Guo, Xu, and Kankanhalli]{BCOM}
Guangzhi Wang, Yangyang Guo, Ziwei Xu, and Mohan~S. Kankanhalli.
\newblock Bilateral adaptation for human-object interaction detection with occlusion-robustness.
\newblock In \emph{CVPR}, 2024.

\bibitem[Wang et~al.(2021)Wang, Yap, Ding, Wu, Yuan, and Tan]{SWiG-HOI}
Suchen Wang, Kim{-}Hui Yap, Henghui Ding, Jiyan Wu, Junsong Yuan, and Yap{-}Peng Tan.
\newblock Discovering human interactions with large-vocabulary objects via query and multi-scale detection.
\newblock In \emph{ICCV}, 2021.

\bibitem[Wang et~al.(2022)Wang, Duan, Ding, Tan, Yap, and Yuan]{THID}
Suchen Wang, Yueqi Duan, Henghui Ding, Yap{-}Peng Tan, Kim{-}Hui Yap, and Junsong Yuan.
\newblock Learning transferable human-object interaction detector with natural language supervision.
\newblock In \emph{CVPR}, 2022.

\bibitem[Xi et~al.(2023)Xi, Meng, and Yuan]{VIDEO_ANALYSIS}
Nan Xi, Jingjing Meng, and Junsong Yuan.
\newblock Open set video {HOI} detection from action-centric chain-of-look prompting.
\newblock In \emph{ICCV}, 2023.

\bibitem[Yang et~al.(2024)Yang, Li, Zeng, Zhang, and Zhang]{MP-HOI}
Jie Yang, Bingliang Li, Ailing Zeng, Lei Zhang, and Ruimao Zhang.
\newblock Open-world human-object interaction detection via multi-modal prompts.
\newblock In \emph{CVPR}, 2024.

\bibitem[Yuan et~al.(2023)Yuan, Zhang, Wang, Albanie, Pan, Feng, Jiang, Ni, Zhang, and Zhao]{RLIPv2}
Hangjie Yuan, Shiwei Zhang, Xiang Wang, Samuel Albanie, Yining Pan, Tao Feng, Jianwen Jiang, Dong Ni, Yingya Zhang, and Deli Zhao.
\newblock Rlipv2: Fast scaling of relational language-image pre-training.
\newblock In \emph{ICCV}, 2023.

\bibitem[Zhai et~al.(2022)Zhai, Wang, Mustafa, Steiner, Keysers, Kolesnikov, and Beyer]{LiT}
Xiaohua Zhai, Xiao Wang, Basil Mustafa, Andreas Steiner, Daniel Keysers, Alexander Kolesnikov, and Lucas Beyer.
\newblock Lit: Zero-shot transfer with locked-image text tuning.
\newblock In \emph{CVPR}, 2022.

\bibitem[Zhang et~al.(2021{\natexlab{a}})Zhang, Liao, Liu, Lu, Wang, Gao, and Li]{CDN}
Aixi Zhang, Yue Liao, Si Liu, Miao Lu, Yongliang Wang, Chen Gao, and Xiaobo Li.
\newblock Mining the benefits of two-stage and one-stage {HOI} detection.
\newblock In \emph{NeurIPS}, 2021{\natexlab{a}}.

\bibitem[Zhang et~al.(2021{\natexlab{b}})Zhang, Campbell, and Gould]{SCG}
Frederic~Z. Zhang, Dylan Campbell, and Stephen Gould.
\newblock Spatially conditioned graphs for detecting human-object interactions.
\newblock In \emph{ICCV}, 2021{\natexlab{b}}.

\bibitem[Zhang et~al.(2022)Zhang, Campbell, and Gould]{UPT}
Frederic~Z. Zhang, Dylan Campbell, and Stephen Gould.
\newblock Efficient two-stage detection of human-object interactions with a novel unary-pairwise transformer.
\newblock In \emph{CVPR}, 2022.

\bibitem[Zhong et~al.(2021{\natexlab{a}})Zhong, Ding, Qu, and Tao]{PD-NET}
Xubin Zhong, Changxing Ding, Xian Qu, and Dacheng Tao.
\newblock Polysemy deciphering network for robust human-object interaction detection.
\newblock \emph{IJCV}, 2021{\natexlab{a}}.

\bibitem[Zhong et~al.(2021{\natexlab{b}})Zhong, Qu, Ding, and Tao]{GGNET}
Xubin Zhong, Xian Qu, Changxing Ding, and Dacheng Tao.
\newblock Glance and gaze: Inferring action-aware points for one-stage human-object interaction detection.
\newblock In \emph{CVPR}, 2021{\natexlab{b}}.

\bibitem[Zhou et~al.(2020)Zhou, Wang, Qi, Ling, and Shen]{CHOIR}
Tianfei Zhou, Wenguan Wang, Siyuan Qi, Haibin Ling, and Jianbing Shen.
\newblock Cascaded human-object interaction recognition.
\newblock In \emph{CVPR}, 2020.

\end{thebibliography}

}

\newpage
\appendix

\maketitlesupplementary

\begin{figure*}[ht]
  \centering
  \includegraphics[width=.95\linewidth]{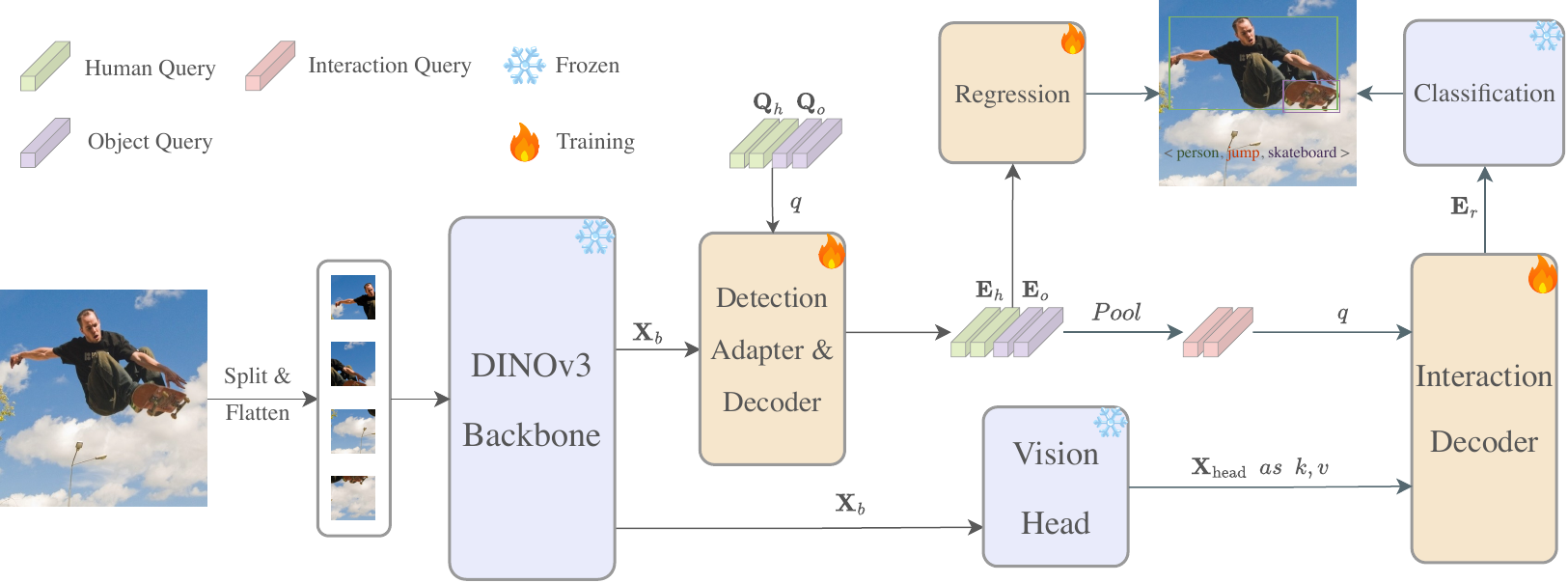}
  \caption{
  Overall architecture of our baseline framework.
  }
  \label{fig:baseline_framework}
\end{figure*}

\begin{figure*}[ht]
  \centering
  \includegraphics[width=.95\linewidth]{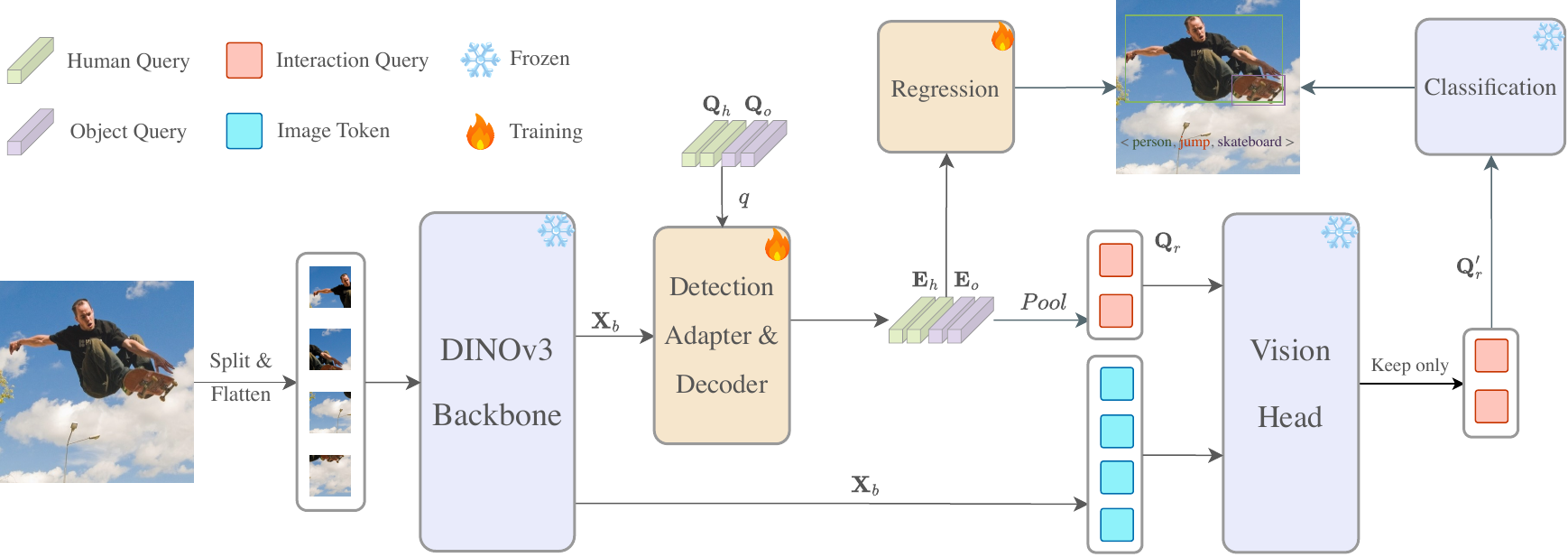}
  \caption{
  Overall architecture of our Semantic Bootstrapping (only) framework.
  }
  \label{fig:semantic_bootstrapping_framework}
\end{figure*}

\section{Training Strategy}
\label{sec:training_strategy}

To ensure fair comparison with prior work, we follow the standard training protocols commonly adopted for each benchmark. 
As discussed in Sec.~\ref{subsec:implementation_details}, \swig~\cite{SWiG-HOI} and \hicodet~\cite{HICO-DET} rely on distinct supervisory setups; therefore, rather than performing dataset-specific tuning, we adhere to the established training practices used in prior work.

Specifically, we follow THID~\cite{THID} for \swig and GEN-VLKT~\cite{GEN-VLKT} for \hicodet:

\begin{itemize}
  \item \textbf{\swig:} we adopt most of THID’s default configuration. The loss weights for bounding-box L1, bounding-box GIoU, interaction classification, and confidence prediction are set to $5.0$, $2.0$, $5.0$, and $10.0$, respectively. 
  Training is conducted for $100$ epochs with a base learning rate of $1\times10^{-4}$, decayed by a factor of $10$ at epochs $60$ and $90$.
  For data augmentation, we follow the commonly used small-scale multi-resolution strategy ($224$-$320$), which stabilizes contrastive \hoi training by maintaining rich in-batch negatives. 
  Random horizontal flipping, light color jittering, and interaction-aware random cropping are applied before resizing, followed by ImageNet~\cite{ImageNet} normalization.

  \item \textbf{\hicodet:} we adopt GEN-VLKT’s training scheme. In addition to interaction classification, an auxiliary object-classification loss is used, and thus the confidence loss is removed. 
  A head-level semantic token is further introduced as an additional key-value input for the instance decoder, consistent with prior architectures.  
  The loss weights for bounding-box L1, bounding-box GIoU, object classification, and interaction classification are set to $2.5$, $1.0$, $1.0$, and $2.0$. 
  Training is performed for $60$ epochs with a base learning rate of $1\times10^{-4}$, reduced by a factor of $10$ at epoch $40$. While GEN-VLKT employs a longer $60+30$ schedule, our model converges reliably under a shorter $40+20$ decay pattern. 
  To ensure fair comparison to strong HICO-DET baselines, we adopt the large-scale multi-resolution augmentation ($480$-$800$), together with the same flipping and light color-jittering scheme as in \swig. 
  Interaction-aware cropping may optionally be applied to preserve human-object spatial structure.
\end{itemize}

Overall, supervision on \swig is intrinsically stronger because it has larger label space and is more challenging,  making it a more representative testbed for open-vocabulary \hoi learning.

\section{Clarification on Image Tokens}
Image tokens in this paper contain three types: the \cls token, register tokens~\cite{Registers}, and patch tokens, as introduced in Sec.~\ref{sec:preliminaries}. 
The \cls token captures global image-level semantics; register tokens serve only as auxiliary tokens during the forward pass and are not used by downstream heads; patch tokens encode local visual information and are the only tokens consumed by our instance detector.

During the semantic bootstrapping stage, we follow the input format of the vision head~\cite{dinotxt} while inserting interaction queries before the patch tokens. The token sequence is
\begin{equation*}
\left[\mathrm{\texttt{CLS}},\,
\mathrm{Reg}_1,\dots,\mathrm{Reg}_4,\,
Q_1,\dots,Q_{N_q},\,
P_1,\dots,P_N
\right],
\end{equation*}
where $Q_i$ denotes interaction queries and $P_i$ denotes patch tokens from the backbone.
In the hierarchical refinement stage, the image tokens used as keys and values in cross-attention are
\begin{equation*}
\left[
\mathrm{\texttt{CLS}}',\,
\overline{P'},\,
P_1',\dots,P_N'
\right],
\qquad
\overline{P'}
= \frac{1}{N}\sum_{i=1}^N P_i',
\end{equation*}
where $'$ indicates tokens output by the vision head.
Two considerations motivate this design: 
(1) register tokens are strictly auxiliary and are thus excluded from downstream prediction heads;
(2) text embeddings are aligned with both the \cls token and the mean patch token, reinforcing semantic consistency between text features and global/region-aggregated visual representations.

\section{Text-Encoder-Based Similarity Classifier}
\label{sec:text_encoder_based_similarity_classifier}
For each interaction category $\mathbf{r}_j$, we construct a descriptive sentence ``a photo of a person \textless action+ing\textgreater\ a/an \textless object\textgreater''. 
For example, ``ride horse'' is expressed as ``a photo of a person riding a horse''. This textual description is encoded using the \dinotxt~\cite{dinotxt} text encoder, producing a $2D=2048$-dimensional embedding $\mathbf{e}_t^{(j)}$. 
The first $D=1024$ dimensions correspond to the \cls token, and the remaining $D=1024$ are aligend with the mean-pooled patch tokens. 
To match the dimension, we project the $D=1024$ interaction decoder embeddings $\mathbf{e}_r^{(i)}$ into the same $2D=2048$-dimensional space, enabling direct similarity computation. 
Because the first and second $D$ dimensions of the text embedding capture different semantic levels, we also explored computing similarities separately and combining them via a weighted sum. However, this alternative produced slightly inferior performance compared to the unified projection.

\section{Variant Models}
\label{sec:variant_models}
The architectures of the variant models are presented in Fig.~\ref{fig:baseline_framework} and Fig.~\ref{fig:semantic_bootstrapping_framework}.

\subsection{Baseline Model}
As illustrated in Fig.~\ref{fig:baseline_framework}, the interaction stage of our baseline model follows the overall structure of HOICLIP~\cite{HOICLIP}, employing a late fusion strategy for interaction classification. 
Semantic cues from the VLM are injected into the interaction queries through cross-attention. 
Contrary to HOICLIP, we do not rely on backbone features to assist interaction detection, as validated by the ablation results in Tab.~\ref{tab:swig_ablation_on_method} for ``Late Fusion (Head only)'' and ``Late Fusion (Multi-Scale)''.

Due to the high dimensionality of token representations, the cross-attention output is first projected down to $d=256$. After cross-attention, the resulting interaction embeddings are then projected to match the dimension of the text embedding space.

\subsection{Semantic Bootstrapping Model}
The Semantic Bootstrapping Model is conceptually simple. Compared to the final model, it removes the cross-attention block, as shown in Fig.~\ref{fig:semantic_bootstrapping_framework}. Only the interaction-query outputs from the vision head are retained. 
Our experiments indicate that these tokens alone are already sufficient for interaction classification and exhibit stronger representational quality than the baseline model. 
As demonstrated in the ablation study (Tab.~\ref{tab:swig_ablation_on_architecture}), this design yields consistent and substantial improvements across all metrics.

\section{Failure Cases and Analysis}
We provide a failure-case analysis to clarify where the current model still struggles.

\paragraph{Typical failure categories.}
We observe two representative failure scenarios: crowded scenes and small-object detection.
\begin{itemize}
  \item \textbf{Crowded scenes:} multiple overlapping human--object instances increase assignment ambiguity and can cause missed detections.
  \item \textbf{Small-object detection:} very small targets are sensitive to slight spatial offsets, leading to localization errors in human or object boxes.
\end{itemize}

\begin{figure}[t]
  \centering
  \begin{subfigure}[b]{0.58\columnwidth}
    \centering
    \includegraphics[width=\linewidth]{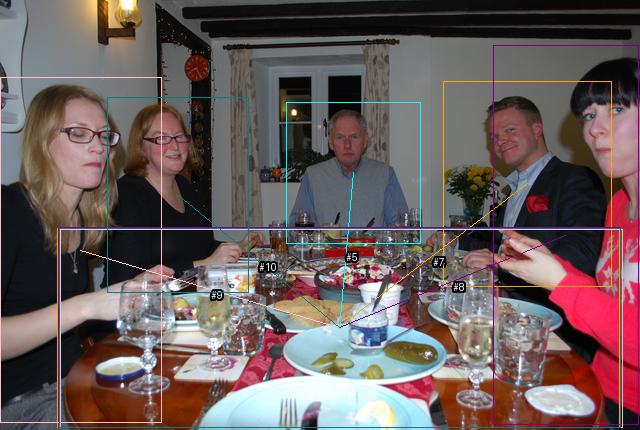}
    \caption{Crowded scene}
    \label{fig:failure_crowded_scene}
  \end{subfigure}
  \hfill
  \begin{subfigure}[b]{0.38\columnwidth}
    \centering
    \includegraphics[width=\linewidth]{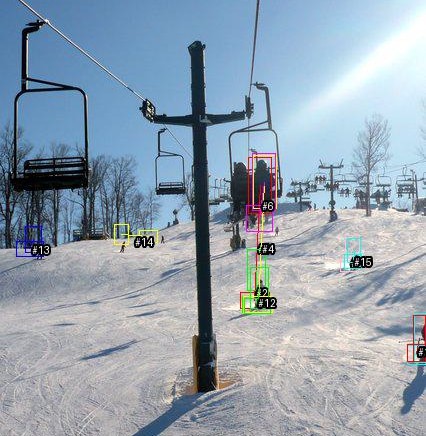}
    \caption{Small-object detection}
    \label{fig:failure_small_object}
  \end{subfigure}
  \caption{Representative failure cases. Left: crowded scene where the detected interactions mainly include \textit{sitting at} and \textit{eating at} a dining table. Right: small-object detection where the detected interactions mainly include \textit{wearing, standing on, holding} a snowboard, and \textit{wearing, carrying, standing on, holding, riding} a skis.}
  \label{fig:failure_cases}
\end{figure}

\paragraph{Error patterns.}
In crowded scenes (e.g., Fig.~\ref{fig:failure_crowded_scene}), while the model successfully detects primary interactions such as \textit{sitting at} and \textit{eating at} a dining table, it is generally robust but still shows occasional misses. For example, a small fork may not be detected even when the main interaction is correctly recognized. In small-object cases (e.g., Fig.~\ref{fig:failure_small_object}), the model identifies complex interaction sets like \textit{wearing, standing on,} and \textit{holding} a snowboard/skis, but both human and object boxes can drift from the true targets. We attribute this mainly to spatial information compression in ViT downsampling, where subtle local offsets of tiny objects become less distinguishable, increasing the failure risk for precise small-object localization.

\section{Training Strategy Comparison}
We compare several training recipes under the same evaluation protocol on \swig (mAP \%).
Beyond the default frozen strategy, we additionally evaluate partial fine-tuning and parameter-efficient adaptation (LoRA \cite{LoRA}).

\paragraph{Template for training recipes.}
For reproducibility, we summarize the practical configurations used in our comparison:
\begin{itemize}
  \item \textbf{Frozen strategy (default):} keep the vision backbone and vision head frozen; train detection adapter, instance decoder, and interaction modules with base learning rate $1\times10^{-4}$.
  \item \textbf{Partial fine-tuning:} unfreeze the vision head and use a reduced learning rate of $2\times10^{-5}$ (i.e., $1/5$ of the base learning rate) to preserve pre-trained semantics during adaptation.
  \item \textbf{LoRA strategy:} apply LoRA on the vision head attention \textit{qkv} input projection and output projection layers, with rank $r=16$ and scaling factor $\alpha=32$; non-LoRA backbone parameters remain frozen.
\end{itemize}

As shown in Tab.~\ref{tab:training_recipe_comparison}, additional training complexity does not necessarily yield meaningful gains; a simple strategy is already effective in practice.

\begin{table}[ht]
\centering
\footnotesize
\setlength{\tabcolsep}{4pt}
\renewcommand{\arraystretch}{1.1}
\caption{Comparison of training recipes on the \swig dataset (mAP \%).}
\label{tab:training_recipe_comparison}
\resizebox{\columnwidth}{!}{\begin{tabular}{lcccc}
\toprule
\textbf{Training Recipe} & \textbf{Unseen} & \textbf{Rare} & \textbf{Seen} & \textbf{Full} \\
\midrule
LoRA fine-tune (all layers) & 18.99 & 24.40 & 29.91 & 24.34 \\
Partial fine-tune (last layer) & 18.63 & 24.37 & 30.01 & 24.26 \\
Ours (frozen) & 19.04 & 24.69 & 30.62 & 24.67 \\
\bottomrule
\end{tabular}}
\end{table}

\section{Pseudo-Code}
\label{sec:pseudo_code}
To illustrate the workflow of our method, we provide a simplified pseudo-code example in Fig.~\ref{fig:pseudocode_listing_corrected}, covering the core computations corresponding to the main equations in Sec.~\ref{sec:method}.

\begin{figure*}[ht]
\begin{lstlisting}[style=pytorchstyle_compact]
import torch.nn as nn
import torch.nn.functional as F

class SL_HOI(nn.Module):
    def __init__(self, backbone, head_model, det_encoder, det_decoder, fusion_decoder):
        super().__init__()
        # Foundation models (parameters are frozen)
        self.backbone = backbone
        self.head_model = head_model

        # Learnable modules
        self.det_encoder = det_encoder
        self.det_decoder = det_decoder
        self.fusion_decoder = fusion_decoder
        self.query_h = nn.Embedding(N_q, d) # Human queries
        self.query_o = nn.Embedding(N_q, d) # Object queries

        # Projection layers for dimension matching
        self.input_proj = nn.Conv2d(D, d, kernel_size=1)
        self.query_proj = nn.Linear(d, D)
        self.cls_proj = nn.Linear(D, D_text)
        self.logit_scale = nn.Parameter(torch.tensor(2.6592))

        # Freeze pre-trained model parameters
        self.freeze_models()

    def freeze_models(self):
        for param in self.backbone.parameters():
            param.requires_grad = False
        for param in self.head_model.parameters():
            param.requires_grad = False

    def forward(self, image, text_features):
        # 1. Feature extraction from frozen backbone 
        # and Interactive instance detection (Eq.`\ref{eq:adapter}` & Eq.`\ref{eq:det_decoder}`)
        with torch.no_grad():
            X_b = self.backbone(image)
        F = self.det_encoder(self.input_proj(X_b))
        E_h, E_o = self.det_decoder(self.query_h.weight, self.query_o.weight, F)

        # 2. Semantic Bootstrapping (Eq.`~\ref{eq:initial_interaction_queries}` & Eq.`~\ref{eq:semantic_bootstrapping}`)
        Q_r = self.query_proj((E_h + E_o) / 2)
        Q_r_prime, X_head = self.head_model(Q_r, X_b)

        # 3. Hierarchical Refinement(Eq.`~\ref{eq:hierarchical_refinement}`)
        E_r = self.fusion_decoder(query=Q_r_prime, key_value=X_head)

        # 4. Open-vocabulary predictions (Eq.`~\ref{eq:open_vocab_classification}`)
        E_r_proj = F.normalize(self.cls_proj(E_r), dim=-1)
        text_features = F.normalize(text_features, dim=-1)
        logits = self.logit_scale.exp() * E_r_proj @ text_features.t()

        return logits
\end{lstlisting}
\caption{An oversimplified PyTorch-style pseudocode of our \ours framework. It illustrates the core logic, including the two-step interaction refinement and the prediction of interaction logits, while omitting many implementation details for clarity.}
\label{fig:pseudocode_listing_corrected}
\end{figure*}

\end{document}